\documentclass[12pt]{article}

\usepackage{color}
\usepackage[T1]{fontenc}    
\usepackage{url}            
\usepackage{booktabs}       
\usepackage{amsfonts}       
\usepackage{nicefrac}       
\usepackage{microtype}      
\usepackage{microtype}
\usepackage{graphicx}
\usepackage{subfigure}
\usepackage{float}
\usepackage{booktabs} 
\usepackage{amsfonts}
\usepackage{bbm}
\usepackage{amssymb,amsmath,amsthm}
\usepackage{amsmath}
\usepackage{amsfonts}
\usepackage{mathrsfs}
\usepackage{amssymb}
\usepackage{amsthm}
\usepackage{wrapfig}
\newtheorem{definition}{Definition}

\usepackage{thmtools}
\usepackage{thm-restate}
\usepackage{hyperref}

\usepackage{cleveref}

\usepackage{graphicx,subfigure}
\usepackage{tikz}
\usepackage{comment}

\usepackage{epstopdf}

\usepackage{libertine}

\usepackage{algorithm}
\usepackage{algorithmic}

\usepackage[numbers]{natbib}
\newdimen\arrowsize
\pgfarrowsdeclare{arcsq}{arcsq}
{
  \arrowsize=0.2pt
  \advance\arrowsize by .5\pgflinewidth
  \pgfarrowsleftextend{-4\arrowsize-.5\pgflinewidth}
  \pgfarrowsrightextend{.5\pgflinewidth}
}
{
  \arrowsize=1.5pt
  \advance\arrowsize by .5\pgflinewidth
  \pgfsetdash{}{0pt} 
  \pgfsetroundjoin   
  \pgfsetroundcap    
  \pgfpathmoveto{\pgfpoint{0\arrowsize}{0\arrowsize}}
  \pgfpatharc{-90}{-140}{4\arrowsize}
  \pgfusepathqstroke
  \pgfpathmoveto{\pgfpointorigin}
  \pgfpatharc{90}{140}{4\arrowsize}
  \pgfusepathqstroke
}

\newtheorem{corollary}{Corollary}

\usepackage{chngcntr}
\usepackage{apptools}
\AtAppendix{\counterwithin{theorem}{section}}
\newtheorem{theorem}{Theorem}
\AtAppendix{\counterwithin{lemma}{section}}
\newtheorem{lemma}{Lemma}
\AtAppendix{\counterwithin{equation}{section}}

\title{An Optimal Transport View on Generalization}

\font\myfont=cmr12 at 13pt

\author{\myfont Jingwei~Zhang\thanks{UBTECH Sydney AI Centre and the School of Information Technologies in the Faculty of Engineering and Information Technologies at The University of Sydney, NSW, 2006, Australia, zjin8228@uni.sydney.edu.au, tongliang.liu@sydney.edu.au, dacheng.tao@sydney.edu.au.} \ \ \   Tongliang~Liu\footnotemark[1] \ \ \    Dacheng~Tao\footnotemark[1]}

\date{}

\begin{document}

\setcitestyle{authoryear,open={},close={}}

\maketitle

\begin{abstract}
We derive upper bounds on the generalization error of learning algorithms based on their \emph{algorithmic transport cost}: the expected Wasserstein distance between the output hypothesis and the output hypothesis conditioned on an input example. The bounds provide a novel approach to study the generalization of learning algorithms from an optimal transport view and impose less constraints on the loss function, such as sub-gaussian or bounded. We further provide several upper bounds on the algorithmic transport cost in terms of total variation distance, relative entropy (or KL-divergence), and VC dimension, thus further bridging optimal transport theory and information theory with statistical learning theory. Moreover, we also study different conditions for loss functions under which the generalization error of a learning algorithm can be upper bounded by different probability metrics between distributions relating to the output hypothesis and/or the input data. Finally, under our established framework, we analyze the generalization in deep learning and conclude that the generalization error in deep neural networks (DNNs) decreases exponentially to zero as the number of layers increases. Our analyses of generalization error in deep learning mainly exploit the hierarchical structure in DNNs and the contraction property of $f$-divergence, which may be of independent interest in analyzing other learning models with hierarchical structure.
 \end{abstract}

\newpage

\section{Introduction}
When designing a learning algorithm,  one fundamental goal is to make the learning algorithm perform well on unseen test data, only with access to a finite number of training data. Formally, we hope that the population risk ( i.e. test error) of the learning algorithm is as small as possible. Unfortunately, minimizing the population risk directly is computationally intractable, because the underlying distribution of the data is unknown. Therefore, to achieve this goal, we separate the population risk as a trade-off sum between empirical risk ( i.e. training error) and the generalization error. The empirical risk measures the extent to which the learning algorithm is consistent with empirical evidence ( i.e. data fitting), while the generalization error quantifies how well the empirical risk can be a valid estimate of the population risk ( i.e. generalization). Thus, one can obtain a hypothesis with minimal population risk by minimizing both the generalization error and empirical risk. Minimizing the empirical risk alone can be realized though empirical risk minimization (\cite{vapnik1999overview}) or its alternatives, for example, the stochastic approximation (\cite{kushner2003stochastic}). However, as the generalization error cannot be minimized directly, it is a common practice to derive a generalization upper bound analytically so that we can study the conditions under which it is guaranteed to be small.

In this paper, we analyze the generalization error of a learning algorithm from an optimal transport perspective. Specifically, our contributions lie in four aspects:
\begin{itemize}
\item We derive an optimal-transport type of generalization error bounds for learning algorithms with Lipschitz continuous loss functions. This result does not put any constraint, for example, the sub-gaussian assumption, on the distribution of input data or the model parameter and applies to unbounded loss functions. 
\item The bound we derive can be related to other probability metrics, e.g., total variation distance, relative entropy, and some notions in classical learning theory, for example, the VC dimension. Therefore, our theory bridges optimal transport theory and information theory with statistical learning theory. 
\item Some other generalization error bounds with different probability metrics are also derived under different assumptions on the loss function. For example, for a learning algorithm with a bounded loss function, a total-variation type generalization error bound can be derived. Via inequalities between probability metrics, the total-variation type generalization error bound can further be bounded by other probability metrics, such as Hellinger distance and $\chi^2$ distance.
\item Under our established framework, we are able to analyze the generalization error in deep learning by exploiting the contraction property of $f$-divergence and the hierarchical structure in DNNs and conclude that the generalization error in DNNs decreases exponentially to zero as the number of layers increases. 
\end{itemize}

The rest of this paper is structured as follows. In Section \ref{section2}, some related works are introduced. Section \ref{section3} formalizes the research problem and gives some basic definitions. In Section \ref{section4}, we derive our main theorem that studies the generalization error of a learning algorithm from an optimal transport perspective. Section \ref{section5} further relates the main theorem to some notions in classical learning theory, for instance, the VC dimension, and derives generalization upper bounds w.r.t. other probability metrics under different conditions for loss functions, which extends the main theorem. In Section \ref{section6}, we analyze the generalization error in deep learning under our established framework and conclude that the key to the non-overfitting puzzle in a DNN is its hierarchical structures.

\section{Related Works} \label{section2}
Our work is related to several different research topics about algorithmic or theoretic aspects of machine learning, summarized below.
\subsection{Generalization in Classical Statistical Learning Theory}
We note that there exists extensive studies on the generalization in classical learning theory. Hence, the references listed here are far from exhaustive.

One central goal in statistical learning theory is to study the condition under which a learning algorithm can generalize. Mathematically, it requires that the generalization error converges asymptotically to zero as the number of training examples goes to infinity.  Traditional ways of charactering the generalization capability of a learning algorithm mainly rely on the complexity of hypothesis class, e.g. VC dimension, Rademacher and Gaussian Complexities, covering number (\cite{vapnik2013nature, bartlett2002rademacher, bartlett2005local, zhou2002covering, zhang2002covering}) or the algorithmic property of the learning algorithm itself, e.g. uniform stability, robustness, algorithmic luckiness (\cite{liu2017algorithmic, shalev2010learnability, bousquet2002stability, xu2012robustness, herbrich2002algorithmic}). These classical learning theory based approaches consider the worst case generalization error over all functions in the hypothesis class and have successfully explained some prevalent learning models, such as the Support Vector Machines (SVMs) for binary classification (\cite{vapnik2013nature}).  Some other approaches are also proposed to analyze the generalization in machine learning, such as PAC-Bayesian approach, Occam's Razor bound, and sample compression approach(\cite{langford2005tutorial, ambroladze2007tighter}).

It is worth mentioning that some works show that these approaches are tightly connected. For example,  \cite{liu2017algorithmic} proves that a higher algorithmic stability implies a smaller hypothesis complexity and \cite{rivasplata2018pac} analyzes the PAC bayesian bounds for stable learning algorithms.
Nevertheless, these approaches are insufficient to explain the generalization of learning models with large hypothesis space, such as deep neural networks (\cite{zhang2016understanding}). Therefore, it is necessary to find a valid approach that can explain why deep learning is attractive in terms of its generalization properties.

\subsection{Information Theoretic Learning and Generalization}
Recently, inspired by an observation in learning theory that learning and information compression are closely related, in the sense that both tasks involve finding identifying patterns or regularities in the training data to re-construct or re-identify future data with similar patterns, some information-theoretic approaches to analyze the learnability and generalization capability of learning algorithms are studied (\cite{nachum2018direct, bassily2018learners, pmlr-v80-alabdulmohsin18a, xu2017information, pmlr-v54-alabdulmohsin17a, pmlr-v51-russo16, bassily2016algorithmic, raginsky2016information, alabdulmohsin2015algorithmic, dwork2015generalization, wang2016average}). Specifically, \cite{pmlr-v51-russo16} show that mutual information the collection of empirical risks and the final output hypothesis can be used to bound the generalization error for learning algorithms with finite hypothesis class. \cite{xu2017information} then extend this result to the case of uncountable infinite hypothesis space and bound the generalization error by the mutual information between the input training set and the output hypothesis, providing a clearer explanation that when the output hypothesis relies less on the training data, the generalization will be better. However, both of these results require that the loss function is sub-gaussian, where the result may not apply to the case when the data distribution is heavy tailed. \cite{alabdulmohsin2015algorithmic} also provides an analysis of stability and generalization from an information-theoretic perspective and shows that the stability of a learning algorithm can be controlled by an information-theoretic notion of algorithmic stability, which is also defined as the mutual information between the input training data and output hypothesis, but the underlying metric defining the mutual information is total variation instead of relative entropy.  However, the analysis of \cite{alabdulmohsin2015algorithmic} restricts to bounded loss functions and countable instance and hypothesis space. In a similar information-theoretic spirit, other information-theoretic notions of stability are proposed: \cite{raginsky2016information} proposes the several information-theoretic notions of stability, such as erasure mutual information, which is an information-theoretic analogous of \emph{replace-one stability} proposed in \cite{shalev2010learnability}; \cite{dwork2015generalization, wang2016average, bassily2016algorithmic} propose notions of stability based on KL divergence, which often arises in the problem of differential privacy.

\subsection{Optimal Transport in Machine Learning}
Optimal transport provides a powerful, flexible, and geometric way to compare probability measures, which is equivalent to measure the Wasserstein distance between two probability distributions (\cite{peyre2017computational})~. Optimal transport has recently drawn attention from the machine learning community, because it is capable of tackling some challenging learning  scenarios such as generative learning (\cite{pmlr-v70-arjovsky17a, gulrajani2017improved}), transfer learning (\cite{courty2017optimal})~, and distributionally robust optimization (\cite{lee2017minimax, gao2016distributionally})~. When comparing different probability distributions, the main advantage of Wasserstein metric over other common probability metrics (such as relative entropy, total variation distance) lies in its good convergence property even when the supports of two probability measures have a a negligible intersection (see example $1$ of \cite{pmlr-v70-arjovsky17a}). To our knowledge, the studies of characterizing the generalization capability of a learning algorithm from an optimal transport perspective are quite limited.

\section{Problem Setup} \label{section3}
We consider the traditional paradigm of statistical learning, where there is an instance space $\mathcal{Z}$, a hypothesis space $\mathcal{W}$, and a nonnegative loss function $\ell: \mathcal{Z}\times\mathcal{W}\to \mathbb{R}^{+} $. The training sample of size $n$ is denoted by $S_n=\{Z_1,\ldots,Z_n\}\in\mathcal{Z}^n$ where each element $Z_i\in\mathcal{Z}, ~i=1,\ldots,n$ is drawn i.i.d. from an unknown underlying distribution $D$. A learning algorithm $\mathcal{A}: \cup_{n=1}^{\infty}\mathcal{Z}^n\to\mathcal{W}$ can be viewed as a (possibly randomized) mapping from the training sample space $\mathcal{Z}^n$ to the hypothesis space $\mathcal{W}$. Following the settings above, the learning algorithm can be uniquely characterized by a Markov kernel $P_{W|S_n}$, which means that given the training sample $S_n$, the learning algorithm picks the output hypothesis $W\in\mathcal{W}$ according the conditional distribution $P_{W|S_n}$. Note that when $P_{W|S_n}$ degenerates to a Dirac Delta distribution, the mapping $\mathcal{A}$ becomes deterministic, which matches the traditional setting of statistical learning, such as Support Vector Machines and linear regression. For any hypothesis $W\in\mathcal{W}$, the population risk is defined as
\begin{equation}
R(W)=\mathbb{E}_{z\sim D}[\ell(z; W)]~.
\end{equation}
The population risk is the performance measure of the hypothesis $W$. Therefore, the goal of learning is to find a hypothesis $W$ with small population risk, either with high probability or by expectation, under the data generating distribution $D$. However, as the underlying population $D$ is unknown, the learning algorithm cannot compute and minimize the population risk $R(W)$ directly. Instead, it can compute the empirical risk of $W$ on the training sample as a proxy, which is defined by
\begin{equation}
R_{S_n}(W)=\mathbb{E}_{z\sim S_n}[\ell(z; h)]=\frac{1}{n}\sum_{i=1}^{n}\ell(z_i; W)
\end{equation}
where $S_n$ is the empirical distribution of the training examples. To evaluate how well the empirical risk can be a valid estimate of the population risk, it defines the generalization error as the difference between the population risk and the empirical risk:
\begin{equation}
G_{S_n}(D, P_{W|S_n}) = R(W)-R_{S_n}(W)~.
\end{equation}
 A small generalization error implies that the empirical risk on the training sample can be a good estimate of the population risk on the underlying unknown population $D$. In this paper, we are interested in expected generalization error, which is defined as
\begin{equation}
G(D, P_{W|S_n}) =\mathbb{E}[R(W)-R_{S_n}(W)]
\end{equation}
where the expectation is taken over the joint distribution of training sample $S_n$ and the hypothesis $W$ (i.e. $P_{S_n,W}=P_{S_n}\times P_{W|S_n}=D^n\times P_{W|S_n}$).

As discussed earlier, the goal of learning is to make the population risk $R(W)$ as small as possible, either in expectation or with high probability. In this paper, we are interested in the expected population risk, which is $\mathbb{E}[R(W)]$. We then have the following decomposition,
\begin{equation}
\mathbb{E}[R(W)]=G(D, P_{W|S_n}) +E[R_{S_n}(W)]~,
\end{equation}
where the first term in the right hand side of the equation controls the generalization and the second term measures the data fitting. To minimize the expected population risk $\mathbb{E}[R(W)]$, we need to minimize both terms. Often, when the training error is small, the learning algorithm tends to fit the training data too well and thus generalizes poorly to unseen test data; when the training error is large, the learning algorithm tends to be insensitive to the training data and therefore has better generalization capabilities towards test data. Thus, a learning algorithm faces a trade-off between minimizing the empirical risk (i.e. data-fitting) and generalization, which is also known as bias-variance trade-off (\cite{mohri2012foundations}) in statistical learning. In what follows, it will be shown how the generalization error can be related to optimal transport. 

\section{Generalization Guarantees via Algorithmic Transport Cost} \label{section4}
We assume that the hypothesis space $\mathcal{W}$ is a Polish space (i.e. a complete separable metric space) with metric $d_{\mathcal{W}}$ and denote by $\mathcal{P}_p(\mathcal{W})$ the space of all Borel probability measures with finite $p$-th ($p\geq1$) moments on $\mathcal{W}$. We can define a family of metrics on the space $\mathcal{P}_p(\mathcal{W})$ with respect to the metric structure $d_{\mathcal{W}}$ on $\mathcal{W}$.
\begin{definition}[Wasserstein Distance] \label{definition1}
For any $p\geq 1$, the $p$-Wasserstein distance between two probability measures $P, Q\in\mathcal{P}_p(\mathcal{W})$ is defined as
\begin{equation}
 \mathbb{W}_p(P, Q) := \inf_{M\in\Gamma(P, Q)}(\mathbb{E}_{(W, W^{\prime})\sim M}[d_{\mathcal{W}}^p(W, W^{\prime})])^{1/p}
 \end{equation}
 where $\Gamma(P, Q)$ denotes the collections of all measures on $W\times W$ with marginals $P$ and $Q$ on the first and second factors respectively. The set $\Gamma(P, Q)$ is also called the set of all couplings of $P$ and $Q$.
\end{definition}
The Wasserstein distance is often used in the problem of optimal transport, which is also called earth mover's distance. Intuitively, for any coupling $\Gamma(P, Q)$ of $W\sim P$ and $W^{\prime}\sim Q$, if each distribution can be viewed as a unit amount of ``dirt'' piled on $\mathcal{W}$, the conditional distribution $P_{W^{\prime}|W}$ can be seen as a randomized policy for moving a unit quantity of dirt from a random location $W$ to another location $W^{\prime}$~. If the cost for moving a unit amount of dirt is $d_{\mathcal{W}}^p(W, W^{\prime})]$, then the optimal cost is identical to the definition of $\mathbb{W}^p_p(P, Q)$~.

In the proof of our main result, we will adopt the Kantorovich dual representation of $1$-Wasserstein metric. It is worth mentioning that as $p$-Wasserstein metric is a monotonic increasing function of $p$ for $p\geq 1$, any upper bound holds for $p=1$ naturally holds for any $p\geq 1$.
\begin{lemma}[\cite{villani2008optimal}]  \label{lemma1}
Let $P, Q$ be two probability measures defined on the same metric space $\mathcal{W}$, i.e., $P, Q\in\mathcal{P}_p(\mathcal{W})$. Then the $1$-Wasserstein distance between $P$ and $Q$ can be represented as
\begin{equation}
 \mathbb{W}_1(P, Q) =\frac{1}{K}\sup_{Lip(f)\leq K} \mathbb{E}_{x\sim P}[f(x)]- \mathbb{E}_{x\sim Q}[f(x)]
\end{equation}
where $Lip(f)$ denotes the Lipschitz constant for $f$.

\end{lemma}
We then define the notion of algorithmic transport cost for a learning algorithm $\mathcal{A}$~.
\begin{definition}[algorithmic transport cost] \label{definition2}
For a learning algorithm $\mathcal{A}:\cup_{n=1}^{\infty} \mathcal{Z}^n\to\mathcal{W}$ with the training set $S\in\mathcal{Z}^n$, the data generating distribution $D$, and the Markov kernel $P_{W|S_n}$, the algorithmic transport cost can be defined as
\begin{equation}
\textit{Opt}(D,P_{W|S_n})=\mathbb{E}_{z\sim D}[\mathbb{W}_1(P_{W}, P_{W|z})]~.
\end{equation}
\end{definition}
\noindent\textbf{Remark 1.} The notion of algorithmic transport cost is defined by the average 1-Wasserstein distance between $W$ and $W|z$, where $z$ is one input example. Intuitively, when one input example $z$ has less contribution on determining $W$, the cost of moving probability mass from $W$ to $W|z$ will become less and thus the generalization will be better. An extreme case is that, when $W$ is independent of $z$, then the generalization error will become zero, since the output hypothesis does not have any correlation with input and therefore has the same performances on both training data and test data.

 Following the definitions and the lemma above, we have our main theorem, which bounds the generalization error of a learning algorithm via algorithmic transport cost.
 \begin{theorem} \label{theorem1}
Assume that the function $W\mapsto\ell(W,z)$ is $K$-Lipschitz continuous for any $z\in\mathcal{Z}$:
\begin{equation}
 |\ell(W,z)-\ell(W^{\prime},z)|\leq K*d_{\mathcal{W}}(W, W^{\prime})~~~for~~any~~W, W^{\prime}\in\mathcal{W}.
 \end{equation}
 The expected generalization error of a learning algorithm $\mathcal{A}$ can be upper bounded by
 \begin{equation}
\mathbb{E}[R(W)-R_{S_n}(W)] \leq K*\textit{Opt}(D, P_{W|S_n})~.
 \end{equation}
 \begin{proof}
 Let $\mathcal{A}: \cup_{n=1}^{\infty} \mathcal{Z}^n\to\mathcal{W}$ be a learning algorithm that has access to a finite set of training examples $S_n=\{Z_i\}_{i=1,\ldots,n}\in\mathcal{Z}^n$ where each element draws i.i.d. from an unknown underlying distribution $D$. Let $W\sim P_{W|S_n}$ be a random variable that stands for the hypothesis output by $\mathcal{A}$ with a Markov kernel $P_{W|S_n}$ and $z\sim S_n$ be a single random training example. By definition, we have
 \begin{eqnarray}\label{flag1}
 &&\mathbb{E}[R(W)-R_{S_n}(W)] \nonumber\\
 &&=\mathbb{E}_{W,S_n}\left[\mathbb{E}_{z\sim D}[\ell(z; W)]-\frac{1}{n}\sum_{i=1}^{n}\ell(z_i; W)\right] \nonumber\\
 &&=\mathbb{E}_{W,S_n}\left[\mathbb{E}_{z\sim D}[\ell(z; W)]-\mathbb{E}_{z\sim S_n}[\ell(z; h)]\right] \nonumber\\
 &&=\mathbb{E}_{W,S_n}\mathbb{E}_{z\sim D}[\ell(z; W)]-\mathbb{E}_{W,S_n}\mathbb{E}_{z\sim S_n}[\ell(z; W)]\nonumber\\
 &&=\mathbb{E}_{W}\mathbb{E}_{z\sim D}[\ell(z; W)]-\mathbb{E}_{W}\mathbb{E}_{S_n|W}\mathbb{E}_{z\sim S_n}[\ell(z; W)]~.
 \end{eqnarray}
 As $z\sim S_n$ is a random training example which is chosen uniformly at random from the training set $S_n$, we have the following relationship between the output hypothesis $W$, the training set $S_n$, and the random training example $z$:
 \begin{equation}
 z\leftarrow S_n \rightarrow W
 \end{equation}
 which means that $z$ and $W$ are conditionally independent of each other when given $S_n$. Therefore, by marginalization, we have
 \begin{eqnarray} \label{flag2}
 &&\mathbb{E}_{S_n|W}\mathbb{P}(z|S_n)=\mathbb{E}_{S_n|W}\mathbb{P}(z|S_n, W)=\mathbb{P}(z|W)~.
 \end{eqnarray}
 Substituting (\ref{flag2}) into (\ref{flag1}) yields
  \begin{eqnarray} \label{flag3}
   &&\mathbb{E}[R(W)-R_{S_n}(W)] \nonumber\\
 &&=\mathbb{E}_{W}\mathbb{E}_{z}[\ell(z; W)]-\mathbb{E}_{W}\mathbb{E}_{z|W}[\ell(z; W)]\nonumber\\
 &&=\mathbb{E}_{z}\mathbb{E}_{W}[\ell(z; W)]-\mathbb{E}_{z}\mathbb{E}_{W|z}[\ell(z; W)]\nonumber\\
 &&=\mathbb{E}_{z}\left[\mathbb{E}_{W}[\ell(z; W)]-\mathbb{E}_{W|z}[\ell(z; W)]\right]\nonumber\\
 &&\leq K*\mathbb{E}_{z}\left[\frac{1}{K}\sup_{Lip(f_z)\leq K} \mathbb{E}_{W\sim P_W}[f_z(W)]- \mathbb{E}_{W\sim P_{W|z}}[f_z(W)]\right] \nonumber\\
 &&=K*\mathbb{E}_{z}\left[\mathbb{W}_1(P_{W}, P_{W|z})\right] \nonumber\\
 &&=K*\textit{Opt}(D, P_{W|S_n})~,
   \end{eqnarray}
  where the inequality follows Lemma \ref{lemma1}, which completes the proof.
 \end{proof}
 \end{theorem}
\textbf{Remark 2.} The derivation of this generalization error bound only requires that the loss function $\ell(W,z)$ is $K$-Lipschitz continuous w.r.t. its first argument $W$. The Lipschitz condition can also be imposed on the second argument $z$ or $(W, z)$. Similar results also hold by slightly modifying the definition of algorithmic transport cost.

\section{Upper Bound via Probability Metrics} \label{section5}
In the previous section, we derive an upper bound on the expected generalization error of a learning algorithm w.r.t. the algorithmic transport cost, which is defined as the expected $1$-Wasserstein distance between the output hypothesis and the output hypothesis conditioned on an input example. In this section, we will first show that the generalization error can also be bounded by other probability metrics via inequalities between probability metrics.

\subsection{Generalization Bound with Total Variation Distance}
In this subsection, we will show that the generalization error can be further upper bounded by total variation distance among distributions relating to input data and output hypothesis. First, we give the definition of total variation distance along with its dual and coupling representations.
\begin{definition}[Total Variation] \label{definition3}
Let $P, Q$ be two probability measures defined on the same metric space $\mathcal{W}$. The total variation distance between $P$ and $Q$ is defined by
\begin{equation}
 \mathbb{TV}(P, Q) := \sup_{A\subset \mathcal{W}} |P(A)-Q(A)|
\end{equation}
where the supremum is over all Borel measurable sets.
\end{definition}
 \begin{lemma}[Dual Representation of Total Variation Distance] \label{lemma2}
Let $P, Q$ be two probability measures defined on the same metric space $\mathcal{W}$. The total variation distance between $P$ and $Q$ is can be represented as
\begin{equation}
\mathbb{TV}(P, Q) =\frac{1}{2F}\sup_{\Vert{f}\Vert_{\infty}\leq F} \mathbb{E}_{x\sim P}[f(x)]- \mathbb{E}_{x\sim Q}[f(x)]~.
\end{equation}
\end{lemma}

\begin{lemma}[Coupling Characterization of Total Variation Distance] \label{lemma3}
Let $P, Q$ be two probability measures defined on the same metric space $\mathcal{W}$. The total variation distance between $P$ and $Q$ has a coupling characterization
\begin{equation}
\mathbb{TV}(P, Q) =\inf_{(X, Y)\sim\Gamma(P, Q)} \mathbb{P}(X\neq Y)~.
\end{equation}
\end{lemma}

With the above definition and lemmas, we derive a generalization upper bound w.r.t. total variation distance.

\begin{theorem} \label{theorem2}
Assume that the hypothesis space $\mathcal{W}$ is bounded, i.e., 
\begin{equation}
 \textit{diam}(\mathcal{W}) := \sup_{w,w^{\prime}\in\mathcal{W}}d_{\mathcal{W}}(w,w^{\prime})<\infty~,
 \end{equation}
 and the function $W\mapsto\ell(W,z)$ is $K$-Lipschitz continuous for any $z\in\mathcal{Z}$~.
  The expected generalization error of a learning algorithm $\mathcal{A}$ can be upper bounded by
 \begin{equation}
\mathbb{E}[R(W)-R_{S_n}(W)] \leq K*\textit{diam}(\mathcal{W})*\mathbb{TV}(P_{W}\times P_{z}, P_{W,z})~.
 \end{equation}
 \begin{proof}
 For any $W, W^{\prime}\in\mathcal{W}$ and $W\sim P_{W}, W^{\prime}\sim P_{W|z}$, we have
 \begin{equation}
 d_{\mathcal{W}}(W, W^{\prime})=\mathbbm{1}_{W\neq W^{\prime}}  d_{\mathcal{W}}(W, W^{\prime})\leq \mathbbm{1}_{W\neq W^{\prime}}*\textit{diam}(\mathcal{W})~.
 \end{equation}
  Taking the infimum of the expected value over $\Gamma(P_{W}, P_{W|z})$ on both sides of the above inequality and following Definition \ref{definition1} and Lemma \ref{lemma3}, we have
   \begin{equation}
 \mathbb{W}_1(P_W,P_{W|z}) \leq \textit{diam}(\mathcal{W})* \mathbb{TV}(P_W,P_{W|z})~.
 \end{equation}
 Following the definition of algorithmic transport cost and Theorem \ref{theorem1}, we obtain
  \begin{equation} \label{equation32}
\mathbb{E}[R(W)-R_{S_n}(W)] \leq K*\textit{diam}(\mathcal{W})*\mathbb{E}_{z\sim D}\left[\mathbb{TV}(P_{W}, P_{W|z})\right]~.
 \end{equation}
 By Definition \ref{definition3}, it follows
 \begin{eqnarray} \label{equation33}
 &&\mathbb{E}_{z\sim D}\left[\mathbb{TV}(P_{W}, P_{W|z})\right] \nonumber \\
 && = \mathbb{E}_{z\sim D}[\sup_{A\subset \mathcal{W}} |P_W(A)-P_{W|z}(A|z)|]\nonumber \\
 && =\int_{\mathcal{Z}}p(z) \sup_{A\subset \mathcal{W}} |P_W(A)-P_{W|z}(A|z)| dz\nonumber \\
 && = \sup_{A\subset \mathcal{W}} \left|P_W(A)\int_{\mathcal{Z}}p(z) dz-\int_{\mathcal{Z}}P_{W,z}(A,z)dz\right|\nonumber \\
 && \leq \sup_{(A,B)\subset \mathcal{W\times Z}}\left|P_W(A)P_z(B)-P_{W,z}(A,B)\right|\nonumber \\
 &&= \mathbb{TV}(P_{W}\times P_{z}, P_{W,z})~.
 \end{eqnarray}
 Combining (\ref{equation33}) and (\ref{equation32}), it completes the proof.
 \end{proof}
\end{theorem}

The above theorem requires that the loss function is Lipschitz continuous and the hypothesis space is bounded. When the loss function is bounded, we can also upper bound the generalization error via total variation distance by exploiting the dual representation of total variation distance. This result is an extension of \cite{alabdulmohsin2015algorithmic}, which requires that both the instance space $\mathcal{Z}$ and the hypothesis $\mathcal{W}$ are countably finite.

\begin{theorem} \label{theorem3}
Assume that the loss function $\ell(W,z)$ is bounded for any $(W, z)$, i.e., 
\begin{equation}
F=\sup_{(W,z)\in\mathcal{W\times Z}}\ell(W,z)<\infty~.
\end{equation}
 The expected generalization error of a learning algorithm $\mathcal{A}$ can be upper bounded by
 \begin{equation}
\mathbb{E}[R(W)-R_{S_n}(W)] \leq 2F*\mathbb{TV}(P_{W}\times P_{z}, P_{W,z})~.
 \end{equation}
 \begin{proof}
 From Equation (\ref{flag3}), the expected generalization error can be rewritten as
 \begin{equation}
 \mathbb{E}_{W}\mathbb{E}_{z}[\ell(z; W)]-\mathbb{E}_{W,z}[\ell(z; W)]~.
 \end{equation}
 By the dual representation of total variation, as shown in Lemma \ref{lemma2}, we have
  \begin{eqnarray}
&& \mathbb{E}_{W}\mathbb{E}_{z}[\ell(z; W)]-\mathbb{E}_{W,z}[\ell(z; W)] \nonumber\\
&& \leq \sup_{\Vert{f}\Vert_{\infty}\leq F} \mathbb{E}_{(W,z)\sim P_{W}\times P_{z}}[f(W,z)]- \mathbb{E}_{(W,z)\sim P_{W,z}}[f(W,z)]\nonumber\\
&&= 2F*\mathbb{TV}(P_{W}\times P_{z}, P_{W,z})
 \end{eqnarray}
 which completes the proof.
 \end{proof}
\end{theorem}

\subsection{Generalization Bound with Mutual Information}
Following the results in the previous section, we can further bound the generalization error via the mutual information between the input training sample $S$ and the output hypothesis $W$. The result in this section is complementary to that of \cite{xu2017information}, in the sense that we do not require that the loss function is sub-gaussian w.r.t. $(W,z)$.
\begin{definition}[Relative Entropy]
Let $P, Q$ be two probability measures defined on the same metric space $\mathcal{W}$.The relative entropy between $P$ and $Q$ is defined by
\begin{equation}
 \mathbb{KL}(P\Vert Q) := \int_{\mathcal{W}}P(x)\log{\frac{P(x)}{Q(x)}}dx~.
\end{equation}
\end{definition}

\begin{definition}[Mutual Information]
Let $X, Y$ be two random variables defined on the same metric space $\mathcal{W}$.The mutual information between $X$ and $Y$ is defined by
\begin{equation}
 I(X, Y) := \mathbb{KL}(P_{X,Y}\Vert P_X\times P_Y) =\int_{\mathcal{W}}P(x,y)\log{\frac{P(x,y)}{P(x)P(y)}}dx~,
\end{equation}
where $P_{X,Y}$ denotes the joint distribution of $(X, Y)$ and $P_X, P_Y$ are corresponding marginal distributions.
\end{definition}

\begin{theorem} \label{theorem4}
Assume that the hypothesis space $\mathcal{W}$ is bounded, i.e., 
\begin{equation}
 \textit{diam}(\mathcal{W}) := \sup_{w,w^{\prime}\in\mathcal{W}}d_{\mathcal{W}}(w,w^{\prime})<\infty~,
 \end{equation}
 and the function $W\mapsto\ell(W,z)$ is $K$-Lipschitz continuous for any $z\in\mathcal{Z}$~.
  The expected generalization error of a learning algorithm $\mathcal{A}$ can be upper bounded by
 \begin{equation}
\mathbb{E}[R(W)-R_{S_n}(W)] \leq K*\textit{diam}(\mathcal{W})*\sqrt{\frac{I(W; S_n)}{2n}}~.
 \end{equation}
 \begin{proof}
Using Pinsker's inequality, we obtain
 \begin{equation}\label{flag4}
 \mathbb{TV}(P_{W}\times P_{z}, P_{W,z})\leq \sqrt{\frac{\mathbb{KL}(P_{W,z}||P_{W}\times P_{z} )}{2}}=\sqrt{\frac{I(W; z)}{2}}~.
 \end{equation}
 As $S=\{z_i\}_{i=1}^{n}\sim D^n$ and each element $z_i$ is drawn i.i.d. from the underlying unknown distribution $D$, we deduce the following inequalities
 \begin{eqnarray}
 &&I(W; S_n) \nonumber \\
 && = I(W; z_1,\ldots,z_n) \nonumber \\
 && = H(z_1,\ldots,z_n)-H_{cond}(z_1,\ldots,z_n|W) \nonumber \\
 && = \sum_{i=1}^{n} H(z_i)-\sum_{i=1}^{n} H_{cond}(z_i|z_{i-1},\ldots,z_1;W) \nonumber \\
 && \geq \sum_{i=1}^{n} H(z_i)-\sum_{i=1}^{n} H_{cond}(z_i|W) \nonumber \\
 && = n I(W; z)~.
 \end{eqnarray}
 It follows naturally 
  \begin{eqnarray} \label{flag5}
 I(W; z)\leq \frac{I(W; S_n)}{n}~.
 \end{eqnarray}
  The proof ends by combining (\ref{flag4}) and (\ref{flag5})~.
 \end{proof}
 \end{theorem}
 
 \subsection{Generalization Bound with Bounded Lipschitz Distance}
 Under assumption that the loss function is both Lipschitz and bounded, we derive the generalization error bound with respect to the Bounded Lipschitz distance.
 \begin{definition}[Bounded Lipschitz Distance] \label{definition6}
Let $P, Q$ be two probability measures defined on the same metric space $\mathcal{W}$.The bounded Lipschitz distance between $P$ and $Q$ is defined by
\begin{equation}
 \mathbb{BL}(P, Q) := \frac{1}{G}\sup_{\Vert f\Vert_{BL}\leq G} \mathbb{E}_{x\sim P}[f(x)]- \mathbb{E}_{x\sim Q}[f(x)]
\end{equation}
where the bounded-Lipschitz norm $\Vert f\Vert_{BL}$ of $f$ is defined by
\begin{equation}
\Vert f\Vert_{BL}=\max\left\{\Vert f\Vert_{\infty},~ \sup_{x\neq y}\frac{|f(x)-f(y)|}{d_{\mathcal{W}}(x,y)} \right\}~.
\end{equation}
\end{definition}
\begin{theorem} \label{theorem5}
Assume that the loss function $\ell(W,z)=\ell_z(W)$ is bounded and Lipschitz continuous continuous with respect to $W$ for any $z\in\mathcal{Z}$, i.e.,  there exists $G$, such that
\begin{equation}
G=\Vert \ell_z\Vert_{BL}<\infty~.
\end{equation}
 The expected generalization error of a learning algorithm $\mathcal{A}$ can be upper bounded by
 \begin{equation}
\mathbb{E}[R(W)-R_{S_n}(W)] \leq G*\mathbb{E}_{z\sim D}[\mathbb{BL}(P_{W}, P_{W|z})]~.
 \end{equation}
 \begin{proof}
 The proof is similar to Theorem \ref{theorem1} where the expected generalization error can be rewritten as
 \begin{equation}
 \mathbb{E}_{W}\mathbb{E}_{z}[\ell(z; W)]-\mathbb{E}_{W,z}[\ell(z; W)]~.
 \end{equation}
By definition \ref{definition6}~, we deduce 
  \begin{eqnarray}
&& \mathbb{E}_{W}\mathbb{E}_{z}[\ell(z; W)]-\mathbb{E}_{W,z}[\ell(z; W)] \nonumber\\
&& \mathbb{E}_{z}\left[\mathbb{E}_{W}[\ell(z; W)]-\mathbb{E}_{W|z}[\ell(z; W)]\right] \nonumber\\
&& \leq  \mathbb{E}_{z}\left[\sup_{\Vert{f_z}\Vert_{BL}\leq G} \mathbb{E}_{W}[f_z(W)]- \mathbb{E}_{W|z}[f_z(W)]\right] \nonumber\\
&&= G*\mathbb{E}_{z}[\mathbb{BL}(P_{W}, P_{W|z})]
 \end{eqnarray}
 which completes the proof.
 \end{proof}
\end{theorem}

\subsection{Relationships to VC-dimension}
We have derived several upper generalization upper bounds with respect to several probability metrics, such as total variation, Wasserstein distance, and relative entropy. In this subsection, we will show how these results can be related to some traditional notions, for example, the VC dimension, that characterize the learnability and generalization capability in statistical learning.

The notion of VC-dimension arises in the problem of binary classification. In this setting, the instance space $\mathcal{Z}=\mathcal{X}\times\mathcal{Y}$, where $\mathcal{X}$ denotes the feature space and $\mathcal{Y}=\{0,1\}$ denotes the label space. The training set is $S_n=\{z_1,\ldots,z_n\}=\{(x_i, y_i)\}_{i=1}^{n}\in\mathcal{X}^n\times\mathcal{Y}^n$. The hypothesis space $\mathcal{W}$ is a class of functions that define the mapping $\mathcal{X}\to \{0, 1\}$~. For any integer $n\geq 0$, we present the definition of the growth function as in \cite{mohri2012foundations}~.

\begin{definition}[Growth Function]
The growth function of a function class $\mathcal{W}$ is defined as
\begin{equation}
\Pi_{\mathcal{W}}(n)=\max_{x_1,\ldots,x_n\in\mathcal{X}} |\{(W(x_1),\ldots,W(X_m)): W\in\mathcal{W}\}|~.
\end{equation}
\end{definition}

In the following theorem, we show how the notion of VC dimension can be related to our previous generalization bounds with probability metrics.
\begin{theorem}
If $\mathcal{W}$ has finite VC-dimension $d$,  the expected generalization error of a learning algorithm $\mathcal{A}$ for binary classification can be upper bounded by
 \begin{equation}
\mathbb{E}[R(W)-R_{S_n}(W)] \leq 2*\mathbb{TV}(P_{W}\times P_{z}, P_{W,z}) \leq\sqrt{\frac{2d\log_{+}(\frac{ne}{d})}{n}}~,
 \end{equation}
 where $\log_{+}(x) := \max\{1,\log x\}$~.
 \begin{proof}
 In the setting of binary classification with hypothesis space $\mathcal{W}$, we have
 \begin{eqnarray}
&& F=\sup_{(W,z)\in\mathcal{W\times Z}}\ell(W,z)\nonumber\\
&&=\sup_{(W,(x,y))\in\mathcal{W\times \mathcal{X}\times\mathcal{Y}}}\ell(W(x),y)\nonumber\\
&&=\max\{\ell(0,1),\ell(1,0),\ell(0,0), \ell(1,1)\} \nonumber\\
&&< \infty.
 \end{eqnarray}
 Without loss of generality, we let $F=1$. Therefore, by Theorem \ref{theorem3}~, we have
 \begin{equation}
\mathbb{E}[R(W)-R_{S_n}(W)] \leq 2*\mathbb{TV}(P_{W}\times P_{z}, P_{W,z})~.
 \end{equation}
 Using Pinsker's inequality and Equation (\ref{flag5}), it follows
 \begin{equation}
 \mathbb{E}[R(W)-R_{S_n}(W)]\leq\sqrt{\frac{2I(S_n; W)}{n}}~.
 \end{equation}
 Denote the collection of empirical risks of the hypothesis in $W$ on $S_n$ by 
 \begin{equation}
 \Lambda_{W}(S_n)=\{R_{S_n}(W)\}_{W\in\mathcal{W}}~.
 \end{equation}
  As the output hypothesis $W$ only depends on the empirical risk of the training data $S_n$,  the following equality holds:
  \begin{eqnarray}
 \sqrt{\frac{2I(S_n; W)}{n}} = \sqrt{\frac{2I(\Lambda_{W}(S_n); W)}{n}}~.
 \end{eqnarray}
 Therefore, we have
  \begin{eqnarray}
&&\sqrt{\frac{2I(\Lambda_{W}(S_n); W)}{n}}\leq \sqrt{\frac{2H(\Lambda_{W}(S_n))}{n}}\nonumber\\
&&\leq  \sqrt{\frac{2\log(\Pi_{\mathcal{W}}(n))}{n}}\leq \sqrt{\frac{2d\log_{+}(\frac{ne}{d})}{n}}
 \end{eqnarray}
 where the last inequality is due to Sauer's lemma
   \begin{eqnarray}
\Pi_{\mathcal{W}}(n) \leq 
\left\{
             \begin{array}{lr}
             2^n, & n<d \\
            \left(\frac{en}{d}\right)^d, & n\geq d  
             \end{array}
\right.~.
 \end{eqnarray}
 \end{proof}
\end{theorem}

\subsection{Generalization Bounds with Other Probability Metrics}
In previous sections, we have derived generalization bounds with different probability metrics under different assumptions (e.g. bounded, Lipschitz) of the loss function. By using the relationships between different probability metrics, in this subsection, we can further extend previous results to other probability metrics listed in Table \ref{table1}. We first present precise definitions of these probability metrics.

Let $P, Q$ be two probability measures defined on the same metric space $\mathcal{W}$. We have the following definitions.
\begin{definition}[Prokhorov metric]
The Prokhorov metric between $P$ and $Q$ is defined by
\begin{equation}
 \mathbb{PR}(P\Vert Q) := \inf_{B\subset\mathcal{W}}\{\epsilon>0: P(B)\leq Q(B^{\epsilon})+\epsilon\}
\end{equation}
where $B^{\epsilon} = \{x: \inf_{y\in B}d_{\mathcal{W}}(x,y)\leq \epsilon\}$ and the infimum is over all Borel sets $B$~.

\end{definition}

\begin{definition}[Hellinger distance]
The Hellinger distance between $P$ and $Q$ is defined by
\begin{equation}
 \mathbb{HL}(P\Vert Q) := \left[\int_{\mathcal{W}}\left(\sqrt{P(x)}-\sqrt{Q(x)}\right)^2dx\right]^{1/2}~.
\end{equation}
\end{definition}

\begin{definition}[$\chi^2$ distance]
The $\chi^2$ distance between $P$ and $Q$ is defined by
\begin{equation}
 \chi^2(P\Vert Q) := \int_{\mathcal{W}}\frac{(P(x)-Q(x))^2}{Q(x)}dx~.
\end{equation}
\end{definition}

\begin{figure} 
  \centering
    \includegraphics[width=0.9\textwidth]{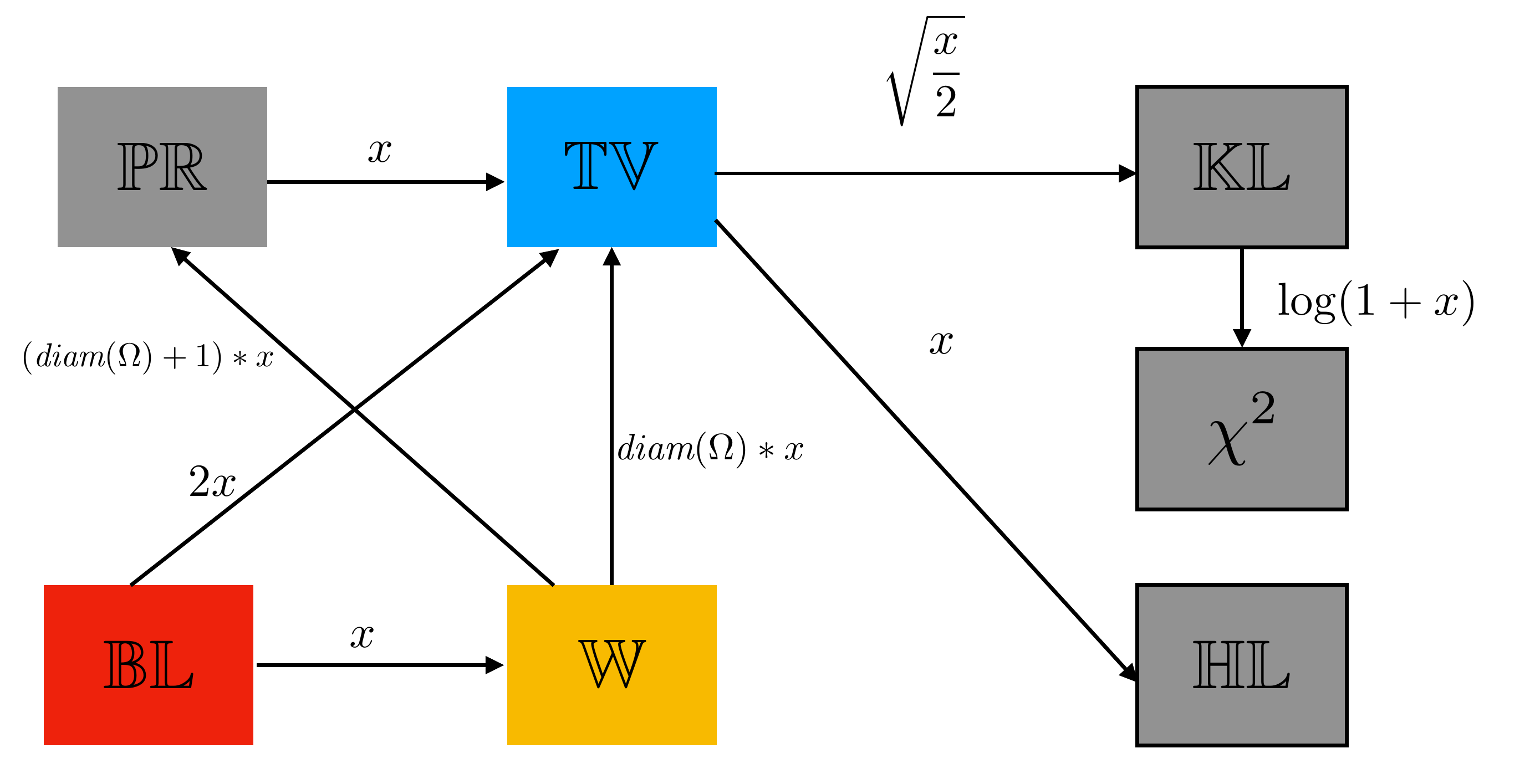}
    \caption{Relationships among different probability metrics on the same measurable space $\Omega$. A directed arrow from metric $\mathbb{A}$ to metric $\mathbb{B}$ annotated by a function $g(x)$ means that $\mathbb{A}(\cdot,\cdot)\leq g(\mathbb{B}(\cdot,\cdot))$~. The symbol $\textit{diam}(\Omega)$ denotes the diameter of the probability space.}
      \label{fig1}
\end{figure}

We present some relationships among different probability metrics in Table \ref{table1}.
\begin{lemma} \label{lemma4}
Let $\Omega$ be any metric space with metric $d_{\Omega}$ and $P, Q$ be two probability measures on $\Omega$. Then the following relationships holds.

$(1)$.~The Wasserstein and Prokhorov metrics satisfy
\begin{equation}
\mathbb{W}(P, Q)\leq (\textit{diam}(\Omega)+1)*\mathbb{PR}(P, Q)
\end{equation}
where $\textit{diam}(\Omega)=\sup_{x, y\in\Omega}d_{\Omega}(x,y)$ denotes the diameter of the probability space.

$(2)$.~ The Bounded Lipschitz distance  and Wasserstein metric satisfy
\begin{equation}
\mathbb{BL}(P, Q)\leq\mathbb{W}(P, Q)~.
\end{equation}

$(3)$.~ The Bounded Lipschitz distance  and total variation distance satisfy
\begin{equation}
\mathbb{BL}(P, Q)\leq 2*\mathbb{TV}(P, Q)~.
\end{equation}

$(4)$. The total variation distance and Hellinger distance satisfy
\begin{equation}
\mathbb{TV}(P, Q)\leq \mathbb{HL}(P, Q)~.
\end{equation}

$(5)$.The relative entropy and $\chi^2$ distance satisfy
\begin{equation}
\mathbb{KL}(P\Vert Q)\leq \log(1+\chi^2(P \Vert Q))\leq\chi^2(P \Vert Q)~.
\end{equation}

\begin{proof}

(1).  This result is due to Theorem 2 in \cite{gibbs2002choosing}.

~~~~~(2) \& (3). These two results are classical facts in probability theory, which can be proved by combining Definition \ref{definition6} with the Kantorovich dual representation of $1$-Wasserstein metric in Lemma \ref{lemma1} and the dual representation of total variation distance in Lemma \ref{lemma2} respectively.

~~~~~(4). See p.35 in \cite{le1969theorie} .

~~~~~(5). This result is proved by \cite{gibbs2002choosing} in Theorem 5.
\end{proof}
\end{lemma}
 The relationships are illustrated in Figure \ref{fig1}. Combining it with results in previous subsections, we have the following generalization bounds for Lipschitz, bounded, bounded-Lipschitz loss functions respectively.

\begin{corollary}[Generalization Bounds for Lipschitz Continuous Loss functions]
Assume that the hypothesis space $\mathcal{W}$ is bounded, i.e., 
\begin{equation}
 \textit{diam}(\mathcal{W}) := \sup_{w,w^{\prime}\in\mathcal{W}}d_{\mathcal{W}}(w,w^{\prime})<\infty~,
 \end{equation}
 and the function $W\mapsto\ell(W,z)$ is $K$-Lipschitz continuous for any $z\in\mathcal{Z}$~. The following generalization bounds holds:
 
\noindent (1). \textbf{Generalization Bound by Prokhorov metric.}
 \begin{equation}
\mathbb{E}[R(W)-R_{S_n}(W)] \leq K*(\textit{diam}(\mathcal{W})+1)*\mathbb{E}_{z\sim D}[\mathbb{PR}(P_{W}, P_{W|z})]~.
 \end{equation}
(2). \textbf{Generalization Bound by Hellinger distance.}
 \begin{equation}
\mathbb{E}[R(W)-R_{S_n}(W)] \leq K*\textit{diam}(\mathcal{W})*\mathbb{HL}(P_{W}\times P_{z}, P_{W,z})~.
 \end{equation}
(3). \textbf{Generalization Bound by $\chi^2$ distance.}
 \begin{equation} 
\mathbb{E}[R(W)-R_{S_n}(W)] \leq K*\textit{diam}(\mathcal{W})*\sqrt{\frac{\log\left(1+\chi^2(P_{W,S_n}||P_{W}\times P_{S_n})\right)}{2n}}~.
 \end{equation}
 
 \begin{proof}
 (1). The result is due to Theorem \ref{theorem1} and (1) of Lemma \ref{lemma4}.

 ~~~~~(2). The result is due to Theorem \ref{theorem2} and (4) of Lemma \ref{lemma4}.
 
  ~~~~~(3). The result is due to Theorem \ref{theorem4} and (5) of Lemma \ref{lemma4}.

 \end{proof}
\end{corollary}

\begin{corollary}[Generalization Bounds for Bounded Loss functions] \label{corollary2}
Assume that the loss function $\ell(W,z)$ is bounded for any $(W, z)$, i.e., 
\begin{equation}
F=\sup_{(W,z)\in\mathcal{W\times Z}}\ell(W,z)<\infty~.
\end{equation}
The following generalization bounds holds:

\noindent (1). \textbf{Generalization Bound by Mutual Information.}
 \begin{equation}
\mathbb{E}[R(W)-R_{S_n}(W)] \leq  F*\sqrt{\frac{2I(S_n; W)}{n}}~.
 \end{equation}
(2). \textbf{Generalization Bound by Hellinger distance.}
 \begin{equation}
\mathbb{E}[R(W)-R_{S_n}(W)] \leq 2*F*\mathbb{HL}(P_{W}\times P_{z}, P_{W,z})~.
 \end{equation}
(3). \textbf{Generalization Bound by $\chi^2$ distance.}
 \begin{equation} 
\mathbb{E}[R(W)-R_{S_n}(W)] \leq F*\sqrt{\frac{2\log\left(1+\chi^2(P_{W,S_n}||P_{W}\times P_{S_n})\right)}{n}}~.
 \end{equation}
 
  \begin{proof}
 (1). The result is by Theorem \ref{theorem3}, Pinsker's inequality, and Equation (\ref{flag5}).

 ~~~~~(2). The result is due to Theorem \ref{theorem3} and (4) of Lemma \ref{lemma4}.
 
  ~~~~~(3). The result is due to (1) of Corollary \ref{corollary2} and (5) of Lemma \ref{lemma4}.

 \end{proof}
\end{corollary}

\begin{corollary}[Generalization Bounds for Bounded and Lipschitz Continuous Loss functions] \label{corollary3}
Assume that the loss function $\ell(W,z)=\ell_z(W)$ is bounded and Lipschitz continuous continuous with respect to $W$ for any $z\in\mathcal{Z}$, i.e.,  there exists $G$, such that
\begin{equation}
G=\Vert \ell_z\Vert_{BL}<\infty~.
\end{equation}
The following generalization bounds holds:

\noindent (1). \textbf{Generalization Bound by algorithmic transport cost.}
 \begin{equation}
\mathbb{E}[R(W)-R_{S_n}(W)] \leq G*\mathbb{E}_{z\sim D}[\mathbb{W}_1(P_{W}, P_{W|z})]~.
 \end{equation}
(2). \textbf{Generalization Bound by total variation distance.}
 \begin{equation}
\mathbb{E}[R(W)-R_{S_n}(W)] \leq 2G*\mathbb{TV}(P_{W}\times P_{z}, P_{W,z})~.
 \end{equation}
(3). \textbf{Generalization Bound by Prokhorov metric.}
 \begin{equation}
\mathbb{E}[R(W)-R_{S_n}(W)] \leq G*(\textit{diam}(\mathcal{W})+1)*\mathbb{E}_{z\sim D}[\mathbb{PR}(P_{W}, P_{W|z})]~.
 \end{equation}
(4). \textbf{Generalization Bound by Mutual Information.}
 \begin{equation}
\mathbb{E}[R(W)-R_{S_n}(W)] \leq  2G*\sqrt{\frac{2I(S_n; W)}{n}}~.
 \end{equation}
(5). \textbf{Generalization Bound by Hellinger distance.}
 \begin{equation}
\mathbb{E}[R(W)-R_{S_n}(W)] \leq 2G*\mathbb{HL}(P_{W}\times P_{z}, P_{W,z})~.
 \end{equation}
(6). \textbf{Generalization Bound by $\chi^2$ distance.}
 \begin{equation} 
\mathbb{E}[R(W)-R_{S_n}(W)] \leq 2G*\sqrt{\frac{2\log\left(1+\chi^2(P_{W,S_n}||P_{W}\times P_{S_n})\right)}{n}}~.
\end{equation}
  \begin{proof}
 (1). The result is due to Theorem \ref{theorem5} and (2) of Lemma \ref{lemma4}. 
 
 ~~~~~(2). The result is due to Theorem \ref{theorem5} and (3) of Lemma \ref{lemma4}.
 
 ~~~~~(3). The result is due to (1) of Corollary \ref{corollary3} and (1) of Lemma \ref{lemma4}.
 
 ~~~~~(4). The result is due to (2) of Corollary \ref{corollary3}, Pinsker's inequality, and Equation (\ref{flag5}).
 
 ~~~~~(5). The result is due to (2) of Corollary \ref{corollary3} and (4) of Lemma \ref{lemma4}.
  
 ~~~~~(6). The result is due to (4) of Corollary \ref{corollary3} and (5) of Lemma \ref{lemma4}.
 
 \end{proof}
\end{corollary}

\begin{table} 
\centering
\begin{tabular}{|c|l|l|l|l|c|l|l|l|l|l|l|l|l|}
\hline
\multicolumn{5}{|c|}{Abbreviation} & \multicolumn{9}{c|}{Metric} \\ \hline
\multicolumn{5}{|c|}{$\mathbb{W}$} & \multicolumn{9}{c|}{Wasserstein Metric}   \\ \hline
\multicolumn{5}{|c|}{$\mathbb{KL}$}             & \multicolumn{9}{c|}{Relative entropy}                     \\ \hline
\multicolumn{5}{|c|}{$\mathbb{TV}$}             & \multicolumn{9}{c|}{Total variation distance}                     \\ \hline
\multicolumn{5}{|c|}{$\mathbb{BL}$}             & \multicolumn{9}{c|}{Bounded Lipschitz distance}                     \\ \hline
\multicolumn{5}{|c|}{$\mathbb{PR}$}             & \multicolumn{9}{c|}{Prokhorov metric}                     \\ \hline
\multicolumn{5}{|c|}{$\mathbb{HL}$}             & \multicolumn{9}{c|}{Hellinger distance}                     \\ \hline
\multicolumn{5}{|c|}{$\chi^2$}             & \multicolumn{9}{c|}{$\chi^2$ distance}                     \\ \hline
\end{tabular} 
\caption{Abbreviations for Metrics} \label{table1}
\end{table}

\section{Application: Explaining Generalization in Deep Learning} \label{section6}
\begin{figure} 
  \centering
    \includegraphics[width=0.9\textwidth]{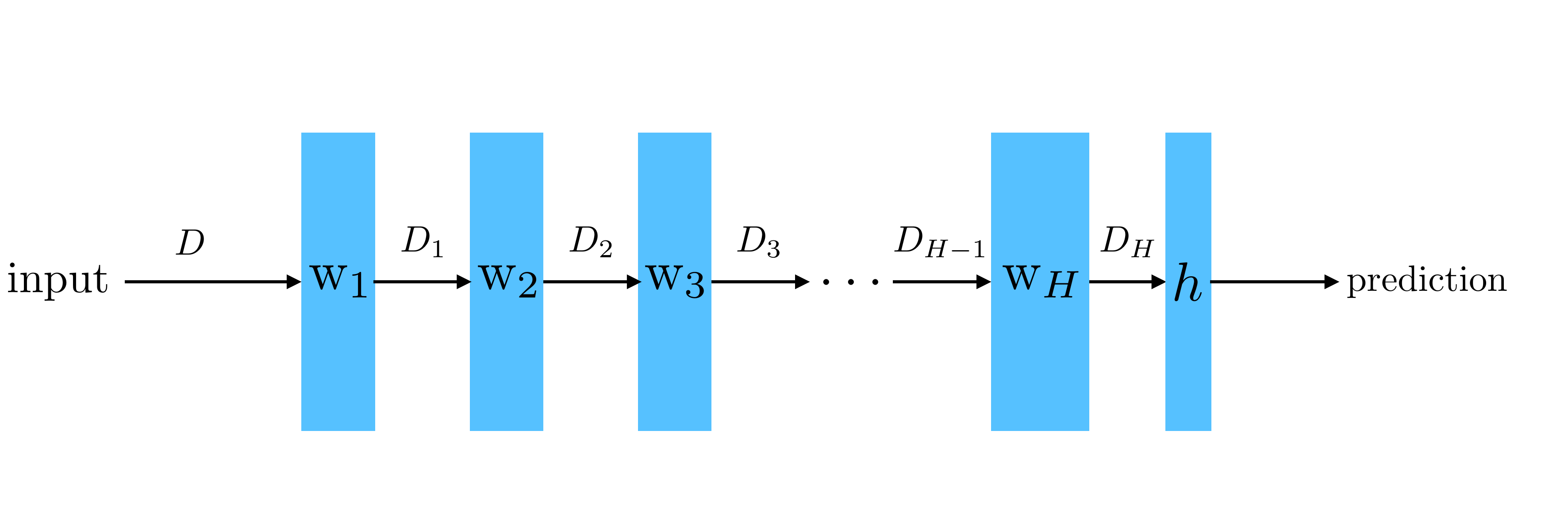}
    \caption{The hierarchical structure of DNNs.} 
      \label{fig2}
\end{figure}

Deep neural networks has shown its attractive generalization capability without explicit regularization even in the heavily over-parametrized regime. Traditional statistical learning theory fails to explain the generalization mystery of deep learning mainly because of the following two reasons:
\begin{itemize}
\item \emph{Worst Case Analysis.}~ The generalization upper bounds derived in traditional statistical learning are based on worst-case analyses  over all functions in the hypothesis space, and thus too loose to bound the generalization error of models with large hypothesis space, such as deep neural networks.
\item \emph{Structure Independence.}~ Traditional statistical learning views a learning model as a whole, ignoring specific structures inside a learning model, such as hierarchical structures in deep neural networks.
\end{itemize} 

In this section, we will explain the non-overfitting puzzle in deep learning via fixing the above two issues arisen in statistical learning. As shown in the proof of Theorem \ref{theorem1}, the analysis of generalization bound via optimal transport only takes the supremum over all Lipchitz functions, and therefore does not rely on worst case analysis as in traditional statistical learning. Besides, in previous sections, we derive generalization error bounds w.r.t. different probability metrics, some of which belong to $f$-divergence (\cite{sason2016f}), such as  total variation distance, relative entropy, Hellinger distance, and $\chi^2$ distance. It is well-known that the strong data processing inequalities (SDPIs) for noisy Markov chains can be applied to $f$-divergence or its related quantities, such as total variation, relative entropy, and mutual information (\cite{polyanskiy2017strong}), which leads to a contraction property. In this subsection, we will use the contraction property for mutual information, as stated in \cite{polyanskiy2017strong}.
\begin{lemma}[Data Processing Inequalities and Strong Data Processing Inequalities for Mutual Information]
Consider a Markov chain $X\rightarrow Y\rightarrow Z$ and the corresponding random mapping $P_{Z|Y}$, by the data processing inequalities, we have $I(X,Z)\leq I(X, Y)$ and the equality holds i.i.f. $X\rightarrow Z\rightarrow Y$ also forms a Markov chain. If the mapping $P_{Z|Y}$ is noisy (that is, we cannot recover $Y$ perfectly from the observed random variable $Z$), then there exists $0\leq\eta<1$, such that
\begin{eqnarray}
I(X,Z)\leq \eta I(X,Y)~.
\end{eqnarray}
\end{lemma}
The above lemma quantifies an intuitive observation that for a Markov chain $X\to Y \to Z$, the noise inside the channel $P_{Z|Y}$ must reduce the information that $Z$ carries about the data $X$ (\cite{ajjanagadde2017lecture}). 
 
A DNN with $H$ hidden layers is illustrated in Figure \ref{fig2}, which conducts $H$ feature transformations sequentially on the input $z\sim D$ and makes predictions on the learned feature $z_L\sim D_L$ by the hypothesis $h$ of the output layer. The output of $i$-th hidden layer is denoted by $z_i\sim D_i$~. The training set is denoted by $S_n=\{z_i\}_{i=1}^{n}$ and the transformed training set at the output of $k$-th hidden layer is denoted by $T_{k_n}=\{z_{k_i}\}_{i=1}^{n}$, where $z_{k_i}$ denotes the $i$-th training sample of the output of $k$-th hidden layer~. The parameter of this DNN is $W=[w_1,\ldots,w_H; h]\in\mathcal{W}=\mathcal{W}_1\times\mathcal{W}_2\times\ldots,\mathcal{W}_H\times\mathcal{H}$ where $\mathcal{W}_i$ is the parameter space of $i$-th hidden layer and $\mathcal{H}$ is the parameter space of the output layer, equipped with metric $d_{\mathcal{H}}$. When given $w_1,\ldots,w_H$, we have the following Markov chain,
\begin{eqnarray}
z\to z_1\to \cdots \to z_H~.
\end{eqnarray} 
Often, as the feature mappings in hidden layers of deep neural networks are noisy (e.g. dropout, noisy SGD) and non-invertible (e.g. convolution, pooling, ReLU activation, non-full column rank in the wight matrix), the channel $z_{i-1}\to z_{i},~ for ~i=1,\ldots,H$ (let $z_0=z$) is often noisy, which will cause a contraction property for the mutual information and we term it a \emph{contraction layer}.  By exploiting the contraction property of hidden layers recursively, we have the following exponential generalization bound w.r.t. the depth $H$ of DNNs, i.e., the generalization error of deep learning will decrease exponentially to zero as we increase the depth of neural networks.

\begin{theorem}[Generalization in Deep Learning] \label{theorem7}
For a DNN with $H$ hidden layers, the input $S_n=\{z_1,\ldots,z_n\}\in\mathcal{Z}^n$, and the output hypothesis $W=[w_1,\ldots,w_H; h]\in\mathcal{W}=\mathcal{W}_1\times\mathcal{W}_2\times\ldots,\mathcal{W}_H\times\mathcal{H}$, assume that the parameter space of the output layer is bounded by $R$, i.e., $R :=\sup_{h, h^{\prime}}d_{\mathcal{H}}(h,h^{\prime})$ and the function $h\mapsto\ell(W,z)=\ell([w_1\ldots,w_H; h],z)$ is $K$-Lipschitz continuous for any $z\in\mathcal{Z}$ and $w_1,\ldots,w_H$:
\begin{equation}
 |\ell([w_1\ldots,w_H; h],z)-\ell([w_1\ldots,w_H; h^{\prime}],z)|\leq K*d_{\mathcal{H}}(h, h^{\prime})~~~for~~any~~h, h^{\prime}\in\mathcal{H}.
 \end{equation}
 Without loss of generality, let all L hidden layers be contraction layers. Then, the expected generalization error can be upper bounded as follows,
\begin{equation}
\mathbb{E}[R(W)-R_{S_n}(W)] \leq  \exp\left(-\frac{H}{2}\log{\frac{1}{\eta}}\right)\sqrt{\frac{K^2R^2I(S_n; W)}{2n}}
 \end{equation}
 where
 \begin{eqnarray} 
\eta = \left( \mathbb{E}_{w_1,\ldots,w_H}\left(\prod_{i=1}^{H} \eta_i\right) \right)^{\frac{1}{H}}<1~.
\end{eqnarray}
denotes the geometric mean of contraction coefficients for all $H$ contraction layers.
 \begin{proof}
 In order to exploit the hierarchical structure in deep neural networks, we have the following decomposition for the expected generalization error, by using the smoothness of the conditional expectation.
 \begin{eqnarray} \label{tag1}
&& \mathbb{E}_{W,S_n}[R(W)-R_{S_n}(W)] 
 \nonumber\\
 &&= \mathbb{E}_{[w_1,\ldots,w_H; h],S_n}[R(W)-R_{S_n}(W)]  \nonumber\\
 && = \mathbb{E}_{w_1,\ldots,w_H}\left[\mathbb{E}_{S_n, h}[R(W)-R_{S_n}(W)|w_1,\ldots,w_H]\right]~.
 \end{eqnarray}
 We then give an upper bound on $\mathbb{E}_{S_n, h}[R(W)-R_{S_n}(W)|w_1,\ldots,w_H]$ in the following lemma.
 
 \begin{lemma} \label{lemma5}
 Under the same condition as in Theorem \ref{theorem7}, the term $\mathbb{E}[R(W)-R_{S_n}(W)|w_1,\ldots,w_H]$ can be upper bounded as
 \begin{equation}
 \mathbb{E}_{S_n, h}[R(W)-R_{S_n}(W)|w_1,\ldots,w_H]\leq \sqrt{\frac{K^2R^2I(h; T_{H_n}|w_1,\ldots,w_H)}{2n}}~.
 \end{equation}
 \begin{proof}
 By definition, we obtain
  \begin{eqnarray}
&& \mathbb{E}_{S_n, h}[R(W)-R_{S_n}(W)|w_1,\ldots,w_H] \nonumber \\
&& = \mathbb{E}_{S_n, h}\left[\mathbb{E}_{z\sim D}[\ell(W, z)]-\frac{1}{n}\sum_{i=1}^{n}\ell(W,z_i)|w_1,\ldots,w_H\right]\nonumber
\\&&= \mathbb{E}_{T_{H_n}, h}\left[\mathbb{E}_{z_H\sim D_H}[\ell(h, z_H)]-\frac{1}{n}\sum_{i=1}^{n}\ell(h,z_{H_i})|w_1,\ldots,w_H\right]\nonumber
\\&&:= \mathbb{E}_{T_{H_n}, h}\left[R(h| w_1,\ldots,w_H)-R_{T_{H_n}}(h| w_1,\ldots,w_H)\right]
 \end{eqnarray}
where $R(h| w_1,\ldots,w_H)$ and $R_{T_{H_n}}(h| w_1,\ldots,w_H)$ denote the conditional expected risk and conditional empirical risk respectively when giving $w_1,\ldots,w_H$. By Theorem $\ref{theorem4}$, the above conditional expected generalization error given $w_1,\ldots,w_H$ can be upper bounded by
\begin{equation}
\mathbb{E}_{T_{H_n}, h}\left[R(h| w_1,\ldots,w_H)-R_{T_{H_n}}(h| w_1,\ldots,w_H)\right]
\leq\sqrt{\frac{K^2R^2I(h; T_{H_n}|w_1,\ldots,w_H)}{2n}}
\end{equation}
which completes the proof. 

It is worth mentioning that the term $I(h; T_{H_n}|w_1,\ldots,w_H)$ refers to the mutual information between the output classifier $h$ and the learned feature $T_{H_n}$ given $w_1,\ldots,w_H$ while the corresponding conditional mutual information is defined as
 \begin{equation}
 I_{cond}(h; T_{H_n}|w_1,\ldots,w_H)=\mathbb{E}_{w_1,\ldots,w_H}[I(h; T_{H_n}|w_1,\ldots,w_H)]~.
 \end{equation}
 \end{proof}
 \end{lemma}

We then upper bound the term $I(h; T_{H_n}|w_1,\ldots,w_H)$ by using SDPIs recursively. When $w_1, \ldots, w_H$ are given, the hierarchical feature mappings in the DNN form a Markov chain,
\begin{eqnarray}
z\to z_1\to \cdots \to z_H~.
\end{eqnarray} 
Therefore, by using SDPIs for contraction hidden layers, we obtain
\begin{eqnarray} \label{tag2}
&& I(h; T_{H_n}|w_1,\ldots,w_H)\leq \eta_H I(h; T_{{H-1}_n}|w_1,\ldots,w_H) \nonumber \\ 
&& \leq \eta_H\eta_{H-1} I(h; T_{{H-2}_n}|w_1,\ldots,w_H)\leq\ldots  \nonumber \\ 
&& \leq \left(\prod_{i=1}^{H} \eta_i\right) I(h; S_n|w_1,\ldots,w_H)
\end{eqnarray} 
Combining (\ref{tag1}), (\ref{tag2}), and Lemma \ref{lemma5}, we obtain,
\begin{eqnarray} \label{tag3}
&& \mathbb{E}_{W,S_n}[R(W)-R_{S_n}(W)]  \nonumber\\
 && = \mathbb{E}_{w_1,\ldots,w_H}\left[\mathbb{E}_{S_n, h}[R(W)-R_{S_n}(W)|w_1,\ldots,w_H]\right] \nonumber\\
&& \leq  \mathbb{E}_{w_1,\ldots,w_H}\left[\sqrt{\frac{K^2R^2I(h; T_{H_n}|w_1,\ldots,w_H)}{2n}}\right]\nonumber\\
&& \leq  \mathbb{E}_{w_1,\ldots,w_H}\left[\sqrt{\prod_{i=1}^{H} \eta_i}\sqrt{\frac{K^2R^2I(h; S_n|w_1,\ldots,w_H)}{2n}}\right] \nonumber\\
&&  \leq \sqrt{ \mathbb{E}_{w_1,\ldots,w_H}\left(\prod_{i=1}^{H} \eta_i\right)}\sqrt{\frac{K^2R^2I_{cond}(h; S_n|w_1,\ldots,w_H)}{2n}}~. 
\end{eqnarray}
Using the fact that condition reduces entropy (\cite{cover2012elements} p.27), we have
\begin{eqnarray} \label{tag4}
&& I_{cond}(h; S_n|w_1,\ldots,w_H) \nonumber \\
&& = H_{cond}(S_n|w_1,\ldots,w_H)- H_{cond}(S_n|h;w_1,\ldots,w_H )\nonumber \\
&&\leq H(S_n) - H_{cond}(S_n|W )\nonumber \\
&&= I(S_n; W)~.
\end{eqnarray}
Substituting (\ref{tag4}) into (\ref{tag3}), we obtain
\begin{eqnarray} 
&& \sqrt{ \mathbb{E}_{w_1,\ldots,w_H}\left(\prod_{i=1}^{H} \eta_i\right)}\sqrt{\frac{K^2R^2I_{cond}(h; S_n|w_1,\ldots,w_H)}{2n}}  \nonumber\\
 && \leq\sqrt{ \mathbb{E}_{w_1,\ldots,w_H}\left(\prod_{i=1}^{H} \eta_i\right)}\sqrt{\frac{K^2R^2I(S_n; W)}{2n}}\nonumber\\
 &&  \leq\sqrt{ \eta^H}\sqrt{\frac{K^2R^2I(S_n; W)}{2n}}  \nonumber\\
 && = \exp\left(-\frac{H}{2}\log{\frac{1}{\eta}}\right)\sqrt{\frac{K^2R^2I(S_n; W)}{2n}}
\end{eqnarray}
where
\begin{eqnarray} 
\eta = \left( \mathbb{E}_{w_1,\ldots,w_H}\left(\prod_{i=1}^{H} \eta_i\right) \right)^{\frac{1}{H}}~.
\end{eqnarray}

 \end{proof}
\end{theorem}

\section{Conclusions and Future Remarks} \label{section7}
In this paper, we derive optimal-transport type of generalization bounds for learning algorithms with Lipschitz loss functions. We further extend the main result in two ways: (1). by leveraging the relationships among different probability metrics and hypothesis complexities, we can also bound the generalization error w.r.t. other probability metrics and hypothesis complexities, such as total variation, relative entropy, and VC dimension; (2). under different constraints on the loss function, we also derive generalization bounds based on various probability metrics, such as total variation distance for bounded loss functions.  Finally, we explain the generalization in deep learning under our proposed framework and conclude that the hierarchical structure is the key to the generalization in DNNs. Our results can naturally lead to several extensions, summarized as follows,
\begin{itemize}
\item \emph{From Generalization by Expectation to Generalization with High Probability}. Our established results consider the expected generalization error for learning algorithms. It is nature to extend our results to the case of generalization with high probability by exploiting concentration inequalities.
\item \emph{Algorithm Design}.  It is necessary to design a learning algorithm that is able to leverage a right balance between data fitting and generalization. One natural way is to contain our established generalization upper bound to the objective function via regularization and thus the trade-off can be controlled by the regularization coefficient.
\item \emph{Computational Exploration}. Most of our generalization error bounds consist of probability metrics between the distributions relating to the training sample and/or the output hypothesis.  However, as the instance space and hypothesis space are often high dimensional, it remains a challenging problem to estimate these quantities defined over different probability metrics such as optimal transport cost.
\end{itemize}

\bibliographystyle{apalike}
\bibliography{arxiv}

\end{document}